\documentclass[letterpaper]{article} %
\usepackage{aaai2026}  %
\usepackage{times}  %
\usepackage{helvet}  %
\usepackage{courier}  %
\usepackage[hyphens]{url}  %
\usepackage{graphicx} %
\usepackage{natbib}  %
\usepackage{caption} %
\usepackage[ruled,vlined]{algorithm2e}
\usepackage{xcolor}
\usepackage[hang,flushmargin]{footmisc}
\usepackage{caption}
\usepackage{float}
\usepackage{makecell}
\usepackage{enumitem}

\usepackage{amsmath}         %
\usepackage{amssymb}         %
\usepackage{booktabs}        %
\usepackage{microtype}       %
\usepackage{epsfig}          %
\usepackage{placeins}        %
\usepackage{color, colortbl} %

\usepackage{tabularx}        %
\usepackage{xstring}         %
\usepackage{multirow}        %
\usepackage{xspace}          %

\usepackage[ruled,vlined]{algorithm2e} %
\usepackage{algorithmicx}
\usepackage{algcompatible}

\usepackage{newfloat}
\usepackage{listings}
\DeclareCaptionStyle{ruled}{labelfont=normalfont,labelsep=colon,strut=off} %
\floatstyle{ruled}
\newfloat{listing}{tb}{lst}{}
\floatname{listing}{Listing}
\nocopyright %

\title{AnomalyControl: Learning Cross-modal Semantic Features for Controllable Anomaly Synthesis}
\author{
    Shidan He\textsuperscript{\rm 1, \rm 2}, Lei Liu\textsuperscript{\rm 3}, Xiujun Shu\textsuperscript{\rm 2}, Bo Wang\textsuperscript{\rm 2}, Yuanhao Feng\textsuperscript{\rm 2}, Shen Zhao\textsuperscript{\rm 1}
}
\affiliations{
    \textsuperscript{\rm 1}Sun Yat-sen University\quad
    \textsuperscript{\rm 2}Tencent, WeChat Pay\quad
    \textsuperscript{\rm 3}Ant Group

}

\usepackage{bibentry}

\begin{document}

\maketitle

\begin{abstract}
Anomaly synthesis is a crucial approach to augment abnormal data for advancing anomaly inspection. Benefiting from large-scale pre-training, existing text-to-image anomaly synthesis methods predominantly focus on textual information or coarse-aligned visual features to guide the entire generation process. However, these methods often lack sufficient descriptors to capture the complicated characteristics of realistic anomalies (\textit{e.g.}, the fine-grained visual pattern of anomalies), limiting the realism and generalization of the generation process. To this end, we propose a novel anomaly synthesis framework called AnomalyControl to learn cross-modal semantic features as guidance signals, which could encode the generalized anomaly cues from text-image reference prompts and improve the realism of synthesized abnormal samples. Specifically, AnomalyControl adopts a flexible and non-matching prompt pair (\textit{i.e.}, a text-image reference prompt and a targeted text prompt) \footnote{Non-matching: the text-image reference prompt and targeted text prompt are only required to describe the same anomaly type without needing to match in surface details or materials.}, where a Cross-modal Semantic Modeling (CSM) module is designed to extract cross-modal semantic features from the textual and visual descriptors. Then, an Anomaly-Semantic Enhanced Attention (ASEA) mechanism is formulated to allow CSM to focus on the specific visual patterns of the anomaly, thus enhancing the realism and contextual relevance of the generated anomaly features. Treating cross-modal semantic features as the prior, a Semantic Guided Adapter (SGA) is designed to encode effective guidance signals for the adequate and controllable synthesis process. Extensive experiments indicate that AnomalyControl can achieve state-of-the-art results in anomaly synthesis compared with existing methods while exhibiting superior performance for downstream tasks. 
\end{abstract}

\begin{links}
    \link{Code}{https://github.com/DaniellaHe/AnomalyControl}
\end{links}

\section{Introduction}

\begin{figure*}[t]
\centerline{\includegraphics[width=2\columnwidth]{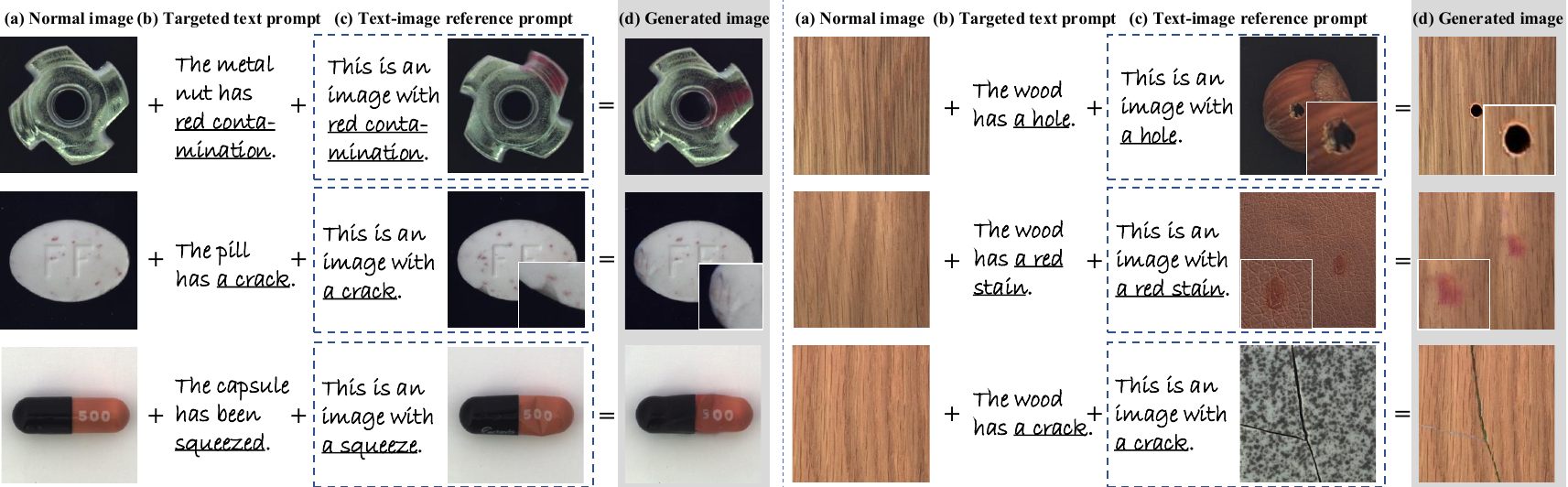}}
\caption{Our method learns cross-modal semantic features as the priors to achieve both \textbf{realism} and \textbf{generalization} for controllable anomaly synthesis. \textbf{Realism:} The targeted text prompt provides a full description for the targeted anomaly image (\textit{e.g.}, color, texture, and shape), while the text-image reference prompt provides a template for the targeted anomaly region (\textit{e.g.}, magnified area). \textbf{Generalization:} The text-image reference can exhibit a different material surface compared with the targeted text prompt (\textit{e.g.}, the right examples).
    }
    \label{fig1}
\end{figure*}

Anomaly inspection plays a crucial role in various fields, from industrial anomaly detection \cite{liu2024deep, realnet} to medical imaging \cite{medical}. One of the most challenging issues is the scarcity of diverse abnormal data due to the resource-intensive procedure for acquiring large and varied datasets. Therefore, anomaly synthesis task \cite{draem, easynet, DiffAug, CDC, DFMGAN, AnoDiff, AnoXFusion, DefGAN, Crop&Paste-SDGAN, anogen} gradually emerged as an advancing technique to improve the data scale of the available abnormal sample.

Many downstream tasks, such as anomaly detection and localization, could benefit from enriching the training dataset using synthetic anomaly samples. 
Recently, text-to-image synthesis techniques \cite{dalle2, sun2024cut} have received more research attention for their outperforming performance,
which can generate visually realistic natural images according to the input text prompt. However, for the anomaly synthesis task, they still suffer from insufficient descriptors of the textual prompt, resulting in unsatisfactory realism for the generated anomalies. This limitation hampers the model's ability to recognize rare or previously unseen defect types.

Motivated by the success of publicly available large text-to-image diffusion models \cite{Imagen, SD, glide, balaji2022ediff,xue2024raphael}, many approaches have been proposed to incorporate extra control signals to assist the diffusion process, which could enrich the targeted text prompt via injecting more prior information. For example, ControlNet \cite{controlnet}, Uni-ControlNet \cite{Uni-controlnet}, and IP-Adapter \cite{Ip-adapter} proposed to apply adapters on stable diffusion models \cite{SD} to provide extra control for generation. The main idea is to learn stylistic information from a reference image (called an image prompt). However, image prompts without a clear focal point generally offer weak alignment signals, limiting the effectiveness of generating high-fidelity and customized images. This limitation is especially pronounced in anomaly synthesis tasks, where anomalies typically occupy only a small region of the image and can easily be overlooked during generation. As a result, the synthesized anomalies often lack sufficient details to match real samples, failing to meet the high precision required for effective anomaly inspection and resulting in a lack of realism in the synthesized samples.

To this end, this work aims to improve the realism of synthesized samples via enhanced controllability of anomaly synthesis. The resulting framework is called AnomalyControl (as shown in Figure \ref{overview}), supporting more flexible and generalizable prompts.  Specifically, the controllability signals come from a non-matching pair, \textit{i.e.}, a text-image reference prompt, and a targeted text prompt. 
The text-image reference prompt includes both textual and visual anomaly descriptors to offer semantic and contextual cues for the synthesis process. To improve realism, we introduce Cross-modal Semantic Modeling (CSM) to capture cross-modal semantic features via multimodal interactions, where the Anomaly-Semantic Enhanced Attention (ASEA) mechanism could extract fine-grained information related to the anomaly region. By taking cross-modal semantic features as the prior, the Semantic Guided Adapter (SGA) could insert effective semantic signals to support an adequate and controllable synthesis process. Table \ref{table1} makes a comparison between AnomalyControl and previous methods, where our method exhibits several advantages in four aspects:
1) \textbf{Realism}: Our model generates highly realistic anomalies by using cross-modal semantic features, closely matching real-world defects.
2) \textbf{Generalization}: Synthesizes diverse anomalies beyond the training distribution, improving adaptability to unseen cases.
3) \textbf{Plug-and-Play}: 
The Semantic Guided Adapter (SGA) allows flexible anomaly synthesis without retraining the diffusion model, making it efficient and scalable.
4) \textbf{Controllability}: AnomalyControl provides precise control over anomaly types and placements using text-image reference and targeted text prompts.

In summary, our contributions are as follows:

(1) A novel framework called AnomalyControl is proposed to achieve controllable anomaly synthesis in a flexible manner, which can generate realistic and generalized anomaly samples guided by a non-matching pair. 

(2) We design a Cross-modal Semantic Modeling (CSM) module to extract cross-modal semantic features from the textual and visual anomaly descriptors, where an Anomaly-Semantic Enhanced Attention (ASEA) mechanism can encode fine-grained visual details into cross-modal semantic features via multimodal interactions, which could further improve the realism of generated anomalies.

(3) We formulate a Semantic Guided Adapter (SGA) allowing cross-modal semantic features to pass through the generation procedure (\textit{i.e.}, diffusion process), providing adequate and controllable synthesis instructions to generate high-quality anomalies.

(4) Extensive experiments demonstrate that our method can generate realistic and generalized anomalies based on different control signals (as shown in Figure \ref{fig1}) and significantly enhance anomaly inspection performance.

\begin{table*}[h]
\centering

\resizebox{\linewidth}{!}{
\begin{tabular}{llccccc}
\hline
\textbf{Method} & \textbf{Venue} & \textbf{Task} & \textbf{Realism} & \textbf{Generalization} & \textbf{Plug-and-Play} & \textbf{Controllability} \\ \hline
DRAEM \cite{draem}           & ICCV'21      & Anomaly Synthesis      &                &                  & \checkmark             &                          \\ 
DefGAN  \cite{DefGAN}        & WACV'21   & Anomaly Synthesis      & \checkmark     &                  &                         &                          \\ 
DFMGAN  \cite{DFMGAN}        & AAAI'23      & Anomaly Synthesis      & \checkmark     &                  &                         &                          \\ 
AnoDiff  \cite{AnoDiff}    & AAAI'24      & Anomaly Synthesis   & \checkmark     &                  &                         &                          \\ 
ControlNet   \cite{controlnet}   & ICCV'23     & Image Generation &                & \checkmark       & \checkmark             & \checkmark               \\ 
Uni-ControlNet \cite{Uni-controlnet}     & NeurIPS'23     & Image Generation &                & \checkmark       & \checkmark             & \checkmark               \\ 
IP-Adapter   \cite{Ip-adapter}   & Arxiv'23     & Image Generation     &                & \checkmark       & \checkmark             & \checkmark               \\ \hline
AnomalyControl  &               & Anomaly Synthesis      & \checkmark     & \checkmark       & \checkmark             & \checkmark               \\ \hline
\end{tabular}}

\caption{Comparisons with previous methods. \textbf{Realism} refers to the ability to generate realistic anomaly images with complex and fine-grained details. \textbf{Generalization} indicates the capability to transfer anomaly features across different objects and surfaces, allowing the model to generate diverse anomalies beyond the pre-training knowledge. \textbf{Our AnomalyControl can perform well for realism and generalization, along with plug-and-play flexibility and controllability.}
}

\label{table1}
\end{table*}

\section{Related Work}

\textbf{Anomaly Synthesis.} Anomaly synthesis has become an essential technique to support anomaly detection systems, especially in scenarios where real defect data is scarce. Existing methods can be broadly divided into two main categories.
First, crop-and-paste anomaly augmentation methods \cite{draem, easynet} combine normal images with abnormal patterns sourced from small abnormal samples or external textures. While these methods are computationally efficient and straightforward to implement, they often lack realism and generalization, limiting their ability to represent the complex characteristics of anomalies. 
Second, generative model-based methods \cite{DiffAug, CDC, DFMGAN, AnoDiff, AnoXFusion, anogen, DefGAN, Crop&Paste-SDGAN}, such as GANs and diffusion models, generate anomalies from scratch or by fine-tuning pre-trained models. Although these methods typically perform well for in-distribution anomalies, they require sufficient sample generalization and frequently struggle to produce out-of-distribution anomalies, which limits their ability to generalize to unseen defect types.

\textbf{Text-to-Image Diffusion Models.} Since fine-tuning large pre-trained models is often inefficient, using adapters has become a more efficient alternative. 
Adapters introduce a small number of trainable parameters into the original model while freezing the pre-trained weights, significantly reducing training costs.
With the success of large text-to-image diffusion models like DALL-E 2 \cite{dalle2}, Imagen \cite{Imagen}, and Stable Diffusion \cite{SD}, significant progress has been made in image generation. However, writing complex text prompts remains a challenge, and pure text often fails to fully express complex scenes or concepts. To address this, methods such as ControlNet \cite{controlnet} and Uni-ControlNet \cite{Uni-controlnet} have introduced lightweight adapters in pre-trained Stable Diffusion models, enabling more precise control by incorporating additional prompts, such as structural or image-based control.
IP-Adapter \cite{Ip-adapter} further optimizes this approach by introducing a decoupled cross-attention mechanism, improving the effectiveness of image prompting, and even achieving performance comparable to fine-tuning the entire model. 
However, image prompts without a clear focus often provide weak alignment signals, making it difficult to generate high-fidelity, customized images, especially in anomaly synthesis, where anomalies are small and easily overlooked.

\section{Methods}

\textbf{Preliminaries.} 
Our method is built upon Stable Diffusion \cite{SD}, which efficiently performs the diffusion process in a low-dimensional latent space instead of pixel space using an auto-encoder \cite{AE}. Specifically, given an input \( x_i \in \mathbb{R}^{H \times W \times 3} \), the encoder maps it to a latent representation \( z_0 = \xi(x_i) \), where \( z_0 \in \mathbb{R}^{h \times w \times c} \) with \( H/h = W/w \) as the downsampling factor and \( c \) as the number of latent dimensions. The diffusion process utilizes a denoising UNet \cite{unet} to gradually refine a noisy latent \( z_t \), conditioned on the current timestep \( T \) and a textual prompt embedding \( C \). \( C \) represents the embedding of the targeted text prompt, generated by a pre-trained CLIP \cite{clip} text encoder. The training objective is defined as:
\begin{equation}
\mathcal{L} = \mathbb{E}_{z_t, t, C, \epsilon \sim \mathcal{N}(0, 1)} \left[ \| \epsilon - \epsilon_{\theta}(z_t, t, C) \|^2_2 \right],
\label{loss}
\end{equation}
where the goal is to predict and remove the noise \( \epsilon \) added to \( z_t \) at each timestep.

\begin{figure*}[t]
\centerline{\includegraphics[width=2\columnwidth]{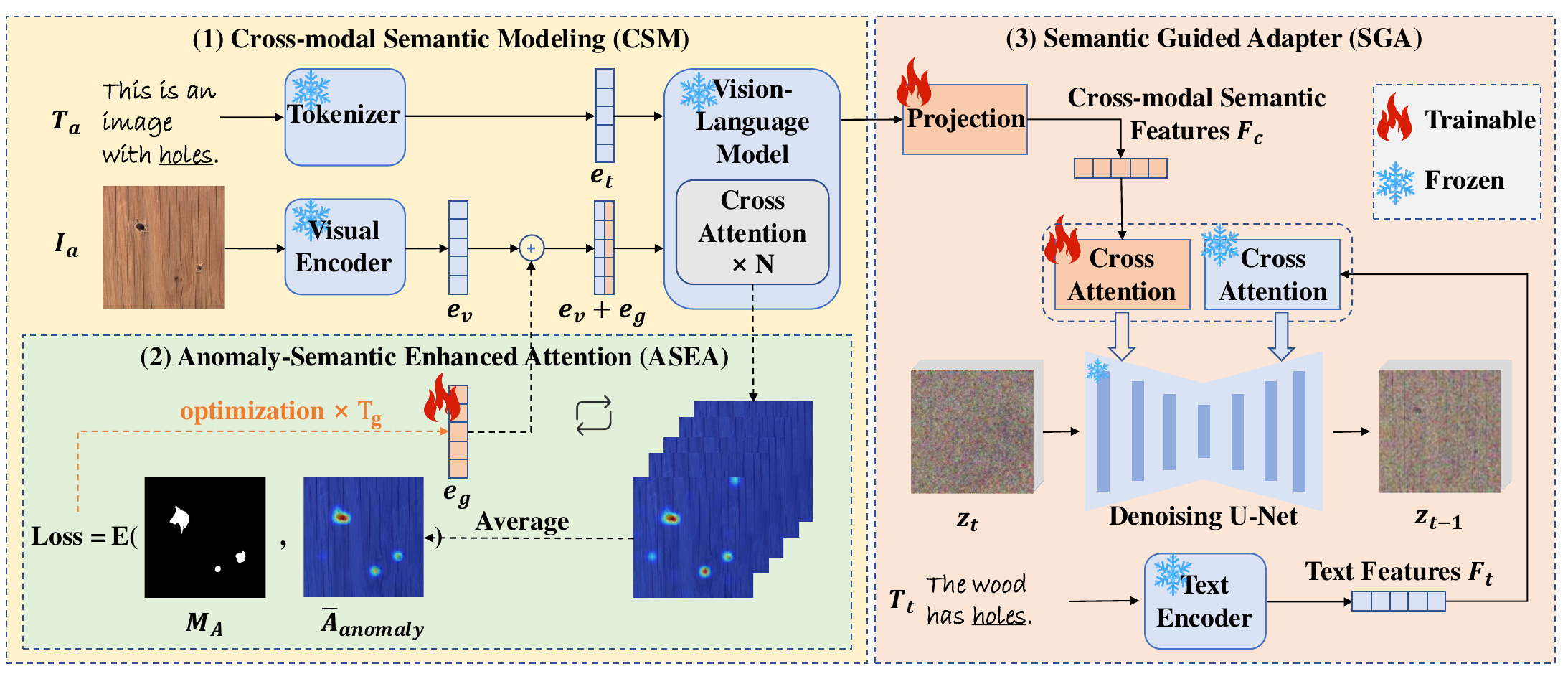}}
\caption{Pipeline of the proposed AnomalyControl. Our framework consists of three modules to enhance controllability in anomaly synthesis. 
(1) \textbf{CSM} integrates both visual and textual anomaly descriptors (\( I_a \), \( T_a \)) to capture precise cross-modal semantic features using a frozen VLM. 
(2) \textbf{ASEA} employs a trainable attention guidance variable \( e_g \) to emphasize the designated anomaly regions, which can capture more accurate semantic information without retraining the frozen VLM. 
(3) \textbf{SGA} utilizes a decoupled cross-attention to support controllable anomaly synthesis, allowing for a flexible and generalized generation. 
}
\label{overview}
\end{figure*}

\textbf{Overall Framework.} The pipeline of AnomalyControl is illustrated in Figure \ref{overview}. 
Our framework consists of two main modules: Cross-modal Semantic Modeling (CSM) and Semantic Guided Adapter (SGA). Given the text-image reference prompts as the input, \textit{i.e.}, a visual anomaly descriptor \( I_a \) and textual anomaly descriptor \( T_a \) as the reference, the CSM module aims to extract fine-grained cross-modal semantic features by integrating both text and image information using a vision-language model (VLM).
An Anomaly-Semantic Enhanced Attention (ASEA) is introduced to refine the attention process in VLM, ensuring that the model can focus on the specific visual patterns of the anomaly, thus enhancing the realism and contextual relevance of the generated anomaly features.
The SGA then introduces these enriched cross-modal semantic features as priors into the diffusion process, allowing flexible control over anomaly characteristics. 
Together, these components ensure that generated anomalies are realistic and contextually aligned with descriptive cues.

\subsection{AnomalyControl}

\subsubsection{Cross-modal Semantic Modeling}
Leveraging the text-image reference prompt, the CSM module targets to encode sufficient cross-modal reference information as control signals for the diffusion process, \textit{i.e.}, consistent anomaly semantics from the text-image prompt. Specifically, the text-image reference prompt provides a visual anomaly descriptor $I_a$ and a textual anomaly descriptor $T_a$. Note that such a text-image reference prompt only focuses on the contextual and descriptive information of the anomaly region, excluding the anomaly-unrelated information, such as background and materials. Such a pattern would allow greater flexibility in anomaly placement. In detail, \( I_a \) defines a specific anomaly pattern to provide a direct visual context reference, ensuring accurate visual control over the generated anomaly. 
\( T_a \) is created by combining an anomaly-specific keyword \( K \) with a template, providing semantic guidance to help the model generate samples that accurately reflect the designated anomaly type and attributes. The anomaly described in the \( T_a \) aligns with the anomaly region in the \( I_a \), while the background and other elements of \( I_a \) remain unconstrained.

To capture effective cross-modal information from the text-image reference, a frozen VLM with cross-attention layers is utilized for its powerful multimodal integration ability. VLM-based CSM could integrate the reference image and text to obtain a cross-modal unified semantic feature, which serves as a prior for the following diffusion process. Natural needs arise from the anomaly-specific regions without losing fine-grained detail to generate realistic and contextually accurate anomalies, which motivates the introduction of the ASEA mechanism in the following section.

\subsubsection{Anomaly-Semantic Enhanced Attention}

ASEA forces the CSM’s attention to focus on designated anomaly regions within the image reference, enabling precise extraction of anomaly-specific features for high-fidelity synthesis. The main pathway is to isolate the anomaly region in the attention map and optimize a trainable attention guidance variable, \( e_g \), which assists the frozen VLM in prioritizing relevant anomaly areas without retraining steps.

The textual anomaly descriptor \( T_a \) is structured as the template of ``\textit{This is an image with [anomaly]}", where the variable \textit{[anomaly]} specifies the anomaly type. \( L_{\text{prefix}} \) represents the length of the fixed prefix ``\textit{This is an image with}". \( L \) denotes the total length of \( T_a \). The anomaly descriptor \textit{[anomaly]} begins at position \( L_{\text{prefix}} + 1 \) and extends to \( L \). 

Firstly, an attention map \( A \) is generated by the VLM’s cross-attention layer, focusing on the feature span \( A_{\text{anomaly}} \) associated with the \([anomaly]\) part of the text prompt. \( A_{\text{anomaly}} \) captures the anomaly-specific attention values:
$
A_{\text{anomaly}} = A[L_{\text{prefix}} + 1 : L, :].
$
By averaging these values across the anomaly section, we compute a mean attention map, \( \bar{A}_{\text{anomaly}} \), which helps stabilize the focus by reducing noise from individual tokens:
\begin{equation}
\bar{A}_{\text{anomaly}} = \frac{1}{L - L_{\text{prefix}}} \sum_{i=L_{\text{prefix}}+1}^{L} A[i, :],
\label{A_anomaly}
\end{equation}

To further refine the attention on anomaly regions, we apply an anomaly region mask \( M_A \) to \( \bar{A}_{\text{anomaly}} \). This mask constrains attention to the spatial area of interest in the image, which is enhanced by introducing the trainable attention guidance variable \( e_g \). The optimization of \( e_g \) is guided by an energy function \cite{layout}, defined as:
\begin{equation}
E(\bar{A}_{\text{anomaly}}, M_A) = \left(1 - \frac{\sum_{i \in M_A} \bar{A}_{\text{anomaly}, i}}{\sum_{i} \bar{A}_{\text{anomaly}, i}}\right)^2,
\label{energy_loss}
\end{equation}
where \( i \in M_A \) refers to indices within the masked region. This function encourages concentration within \( M_A \), prioritizing relevant features in the anomaly area.

Finally, \( e_g \) is optimized iteratively over \( T_g \) steps using gradient descent to minimize the energy function:
\begin{equation}
e_g \leftarrow e_g - \alpha \nabla_{e_g} E(\bar{A}_{\text{anomaly}}, M_A),
\label{e_g}
\end{equation}
where \( \alpha \) is the learning rate. Through this iterative refinement, ASEA aligns the model’s attention precisely with \( M_A \), effectively capturing anomaly-specific details. The challenge here lies in maintaining focused attention on the anomaly regions without disrupting the model’s overall comprehension, a task ASEA addresses by carefully guiding attention with \( e_g \).
\begin{table*}[!t]
\centering

\setlength{\tabcolsep}{3pt} %
\begin{tabular}{l c|ccccccccccccccc|c}
\hline
Method & Metric & bottle & cable & caps & carp & grid & hazel & leath & metal & pill & screw & tile & brush & trans & wood & zipper & Mean \\
\hline
\multirow{2}{*}{DiffAug
}
& IS & 1.59 & 1.72 & 1.34 & 1.19 & 1.96 & 1.67 & 2.07 & 1.58 & 1.53 & 1.10 & 1.93 & 1.33 & 1.34 & 2.05 & 1.30 & 1.58 \\
& IL & 0.03 & 0.07 & 0.03 & 0.06 & 0.06 & 0.05 & 0.06 & 0.29 & 0.05 & 0.10 & 0.09 & 0.06 & 0.05 & 0.30 & 0.05 & 0.09 \\
\hline
\multirow{2}{*}{CDC 
}
& IS & 1.52 & 1.97 & 1.37 & \textbf{1.25} & 1.97 & 1.97 & 1.80 & 1.55 & 1.56 & 1.13 & 2.10 & 1.63 & 1.61 & 2.05 & 1.30 & 1.65 \\
 & IL & 0.04 & 0.19 & 0.06 & 0.03 & 0.07 & 0.05 & 0.07 & 0.04 & 0.06 & 0.11 & 0.12 & 0.06 & 0.13 & 0.03 & 0.05 & 0.07 \\
\hline
\multirow{2}{*}{SDGAN 
}
& IS & 1.57 & 1.89 & 1.49 & 1.18 & 1.95 & 1.85 & 2.04 & 1.45 & 1.61 & 1.17 & \underline{2.53} & 1.78 & \textbf{1.76} & 2.12 & 1.25 & 1.71 \\
& IL & 0.06 & 0.19 & 0.03 & 0.11 & 0.10 & 0.16 & 0.12 & 0.28 & 0.07 & 0.10 & 0.21 & 0.03 & 0.13 & 0.25 & 0.10 & 0.13 \\
\hline
\multirow{2}{*}{DefGAN 
}
& IS & 1.39 & 1.70 & 1.59 & \underline{1.24} & 2.01 & 1.87 & \textbf{2.12} & 1.47 & 1.61 & 1.19 & 2.35 & \textbf{1.85} & 1.47 & 2.19 & 1.25 & 1.69 \\
& IL & 0.07 & 0.22 & 0.04 & 0.12 & 0.12 & 0.19 & 0.14 & 0.30 & 0.10 & 0.12 & 0.22 & 0.03 & 0.13 & 0.29 & 0.10 & 0.15 \\
\hline
\multirow{2}{*}{DFMGAN 
}
& IS & \underline{1.62} & 1.96 & 1.59 & 1.23 & 1.97 & 1.93 & 2.06 & 1.49 & \underline{1.63} & 1.12 & 2.39 & \underline{1.82} & 1.64 & 2.12 & 1.29 & 1.72 \\
 & IL & \underline{0.12} & 0.25 & 0.11 & 0.13 & 0.13 & 0.24 & 0.17 & \underline{0.32} & 0.16 & 0.14 & 0.22 & 0.18 & 0.25 & 0.35 & \textbf{0.27} & 0.20 \\
\hline
\multirow{2}{*}{AnoDiff  
}
& IS & 1.58 & \underline{2.13} & \underline{1.59} & 1.16 & \underline{2.04} & \textbf{2.13} & 1.94 & \underline{1.96} & 1.61 & \textbf{1.28} & \textbf{2.54} & 1.68 & 1.57 & \textbf{2.33} & \underline{1.39} & \underline{1.80} \\

& IL & \textbf{0.19} & \underline{0.41} & \textbf{0.21} & \underline{0.24} & \underline{0.44} & \underline{0.31} & \underline{0.41} & 0.30 & \underline{0.26} & \textbf{0.30} & \textbf{0.55} & \underline{0.21} & \underline{0.34} & \underline{0.37} & 0.25 & \underline{0.32} \\
\hline

\multirow{2}{*}{Ours} 
& IS 
& \textbf{1.63} & \textbf{2.14} & \textbf{1.69} & 1.18 & \textbf{2.26} & \underline{2.12} & \underline{2.08} & \textbf{1.98} & \textbf{1.64} & \underline{1.25} & \textbf{2.54} & 1.80 & \underline{1.66} & \underline{2.20} & \textbf{1.40} & \textbf{1.84} \\

& IL & \textbf{0.19} & \textbf{0.44} & \underline{0.20} & \textbf{0.28} & \textbf{0.47} & \textbf{0.35} & \textbf{0.43} & \textbf{0.34} & \textbf{0.27} & \underline{0.28} & \underline{0.53} & \textbf{0.24} & \textbf{0.39} & \textbf{0.40} & \underline{0.26} & \textbf{0.35}\\
\hline
\end{tabular}

\caption{
Quantitative results of generation evaluated by IS and IC-LPIPS (abbreviated as IL) on the MVTec AD dataset. Bold and underlined values indicate the best and second-best results, respectively.
}

\label{Quantitative generation results}
\end{table*}

\subsubsection{Semantic Guided Adapter}
 Using solely image features as the prior \cite{Ip-adapter, Uni-controlnet}, previous approaches often miss critical details in subtle anomaly regions with low signal-to-noise ratios. Motivated by this, SGA relies on a richer, semantically focused representation, \textit{i.e.}, cross-modal semantic features produced by CSM, which captures fine-grained anomaly details essential for accurate synthesis. Concretely, based on the decoupled cross-attention \cite{Ip-adapter}, SGA introduces an adaptive mechanism to incorporate an additional targeted text prompt \( T_t \) as input.

To facilitate classifier-free guidance \cite{CFG}, SGA employs random dropout during training. This process involves occasionally setting cross-modal semantic features to zero, allowing the model to jointly learn both conditional and unconditional prompts. The enhanced cross-attention mechanism for integrating text and cross-modal semantic features is defined as:
\begin{equation}
Z_{\text{new}} = \mathcal{S}\left(\frac{Q(K)^T}{\sqrt{d}}\right) V + \gamma \cdot \mathcal{S}\left(\frac{Q(K')^T}{\sqrt{d}}\right) V',
\end{equation}
where $\gamma$ is a weight to balance these two terms. $\mathcal{S}$ is the function of Softmax.
\( Q \), \( K \), and \( V \) are the query, key, and value matrices for the attention operation applied to text cross-attention, while \( K^{\prime} \) and \( V^{\prime} \) correspond to cross-modal attention. Given the query features \( Z \) and cross-modal semantic features \( F_c \), the query matrix \( Q \) is defined as \( Q = ZW_q \), with \( K^{\prime} = F_c W_k^{\prime} \) and \( V^{\prime} = F_c W_v^{\prime} \). 
Notably, only \( W_k^{\prime} \) and \( W_v^{\prime} \) are trainable parameters, focusing the adaptation on cross-modal information.
During training, only the parameters of the SGA and \( e_g \) are optimized, while the pre-trained diffusion model and VLM parameters remain frozen. The entire AnomalyControl pipeline is trained on image-text pairs of anomaly images, with a training objective based on the original Stable Diffusion work:
\begin{equation}
\mathcal{L} = \mathbb{E}_{ \epsilon \sim \mathcal{N}(0, 1)} \left[ \| \epsilon - \epsilon_{\theta}(\text{concat}(z_t, M_A, z_m,), t, F_t, F_c) \|^2_2 \right],
\label{loss_new}
\end{equation}
where \(z_m\) represents the masked image latent. This setup allows SGA to integrate anomaly-specific semantic signals effectively, ensuring high-quality and controllable synthesis.

\section{Experiments}

\subsection{Experimental Setup}
\textbf{Datasets.}
We conduct experiments on three datasets, including MVTec AD \cite{bergmann2019mvtec, bergmann2021mvtec}, MPDD \cite{MPDD}, and ViSA \cite{VISA}, to evaluate the performance of our method.
Following the protocol in AnoDiff \cite{AnoDiff}, we adopt the same data split: the first one-third of the anomaly dataset IDs is used for training the anomaly synthesis model, and the remaining two-thirds for testing.
The anomaly synthesis model generates 1,000 pairs of anomaly images and corresponding masks for each anomaly category using normal data. The masks are generated using the mask generation method from AnoDiff \cite{AnoDiff}. 
These pairs of anomaly images and their corresponding masks, along with normal images from the training set, are then used to train downstream anomaly detection models, following the same approach as DRAEM \cite{draem} and AnoDiff \cite{AnoDiff}. 

\textbf{Metrics.} 
Inception Score (IS) \cite{IS} \cite{AnoDiff} is to evaluate the quality and diversity of synthesized anomalies. Intra-cluster Pairwise Learned Perceptual Image Patch Similarity (IC-LPIPS, abbreviated as IL) \cite{ojha2021few} is used to evaluate perceptual similarity and the generalization of synthetic anomalies. For downstream tasks such as anomaly detection and localization, we employed measurements through pixel-level and image-level Area Under the Receiver Operating Characteristic Curve (AUROC), Average Precision (AP), and F1-max scores. 

\textbf{Implementation Details.} AnomalyControl is implemented with the HuggingFace Diffusers library \cite{diffusers2022} and built on the Stable Diffusion 1.5 (SD1.5) model. The VLM utilized in CSM is based on BLIP-2 \cite{blip2}, which provides robust semantic feature extraction for cross-modal inputs. We incorporate a new cross-attention layer into each of SD1.5’s 16 cross-attention layers, setting the cross-modal guidance strength parameter \( \gamma = 1 \) for balanced guidance.
The number of guidance steps \( T_g \) in ASEA is set to 3.
Additional details regarding the implementation are provided in the Appendix.

\begin{figure*}[h]
\centerline{\includegraphics[width=1.9\columnwidth]{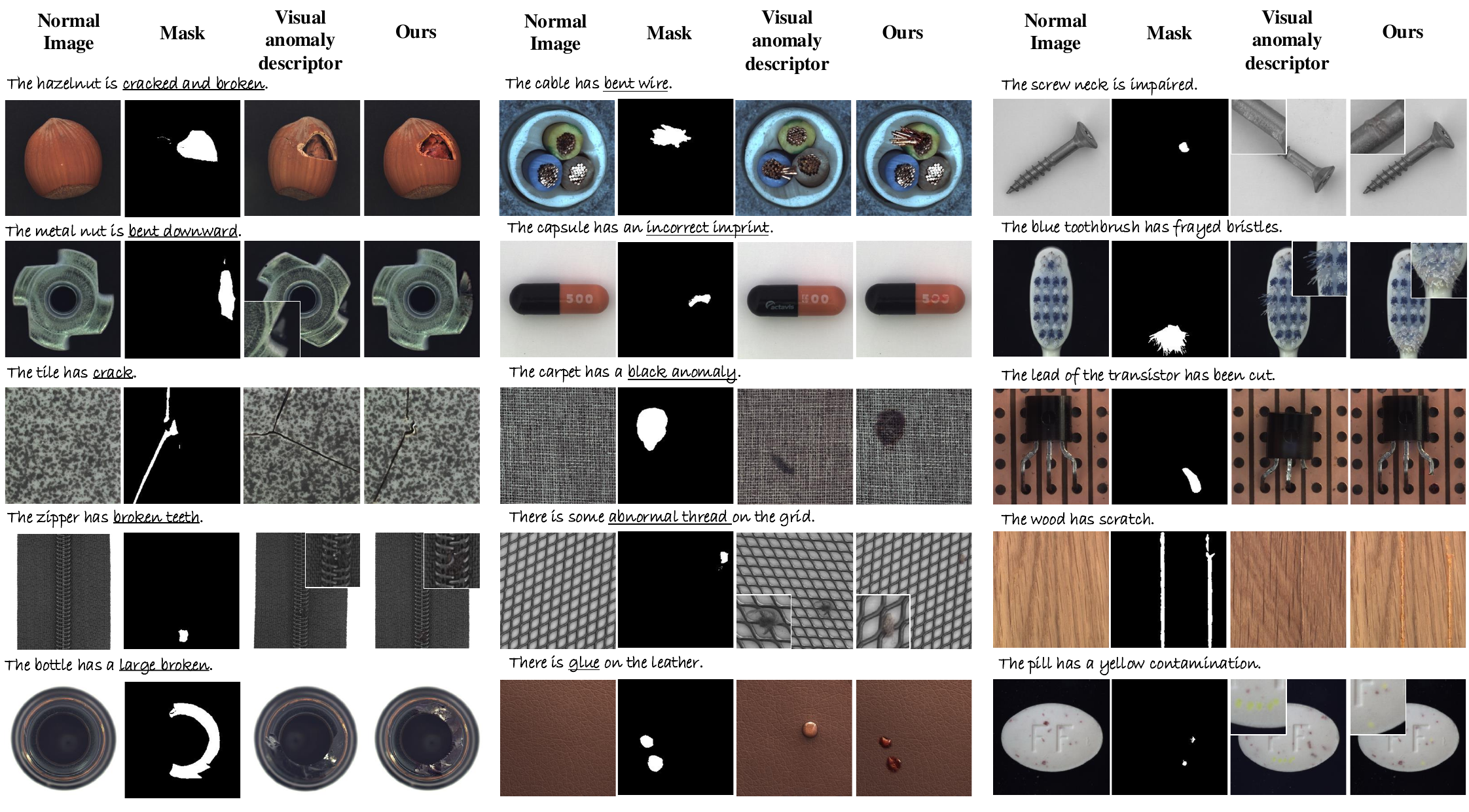}}
\caption{
Display of anomaly synthesis on 15 objects from the MVTec AD dataset. The first column shows the normal image, the second column represents the mask area where anomalies need to be generated, the third column contains the visual anomaly descriptor providing visual guidance, and the fourth column displays the results generated by our method.  
Unlike simply copying the anomaly shown in the visual anomaly descriptor, our model predicts what the anomaly would look like in the given region indicated by the mask, considering the material and texture of the original normal image. The generated anomalies are consistent with the context of the image and are realistic in their appearance.
Zoom in for better clarity and details.
}
\label{supp_ad_1}
\end{figure*}

\begin{figure}[!t]
\centerline{
\includegraphics[width=0.6\columnwidth]{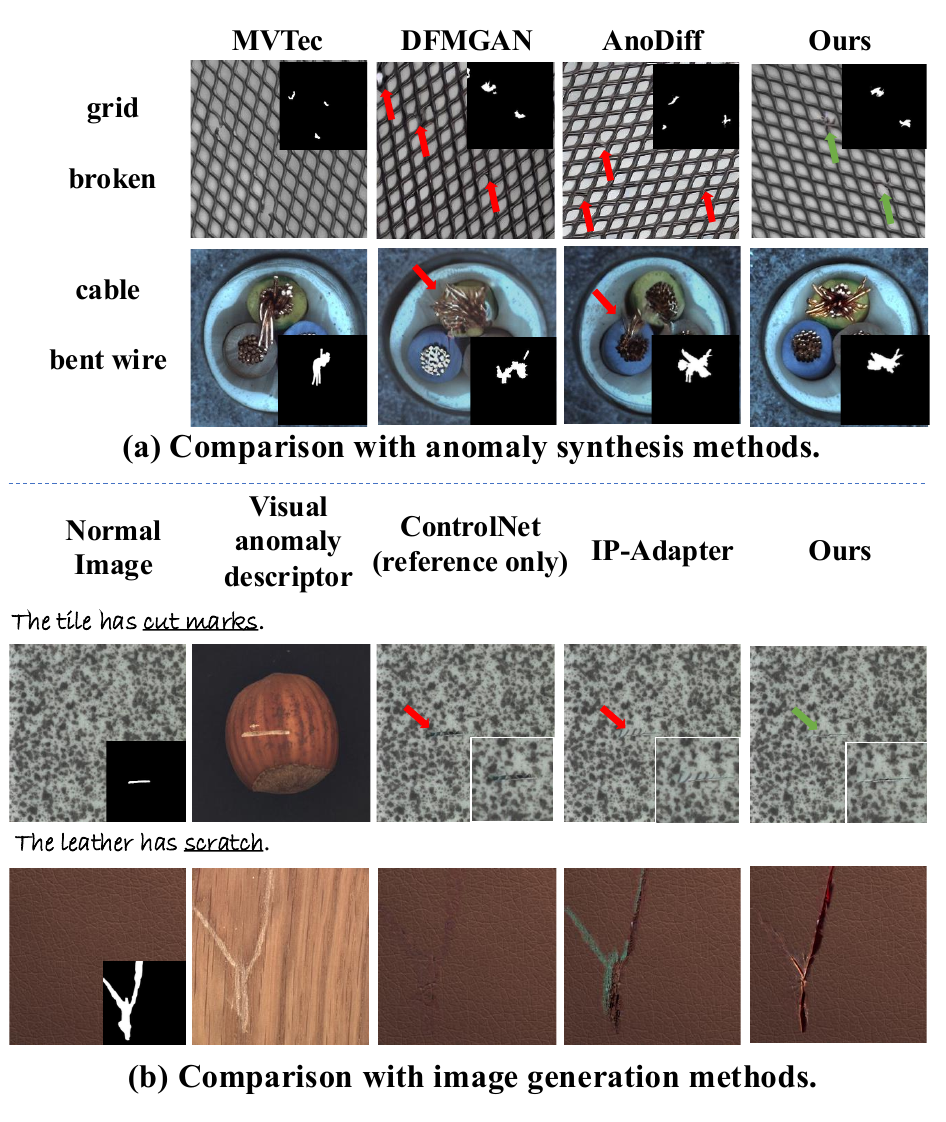}
}
\caption{
Qualitative Evaluation. Our method could exhibit large improvements in realism and generalization compared with both anomaly synthesis and image generation methods. For example, in the grid and cable examples, our method produces the clearest and most realistic anomalies, closely matching the texture and material of the original image.}
\label{QE}
\end{figure}

\begin{figure}[!t]
\centerline{\includegraphics[width=\columnwidth]{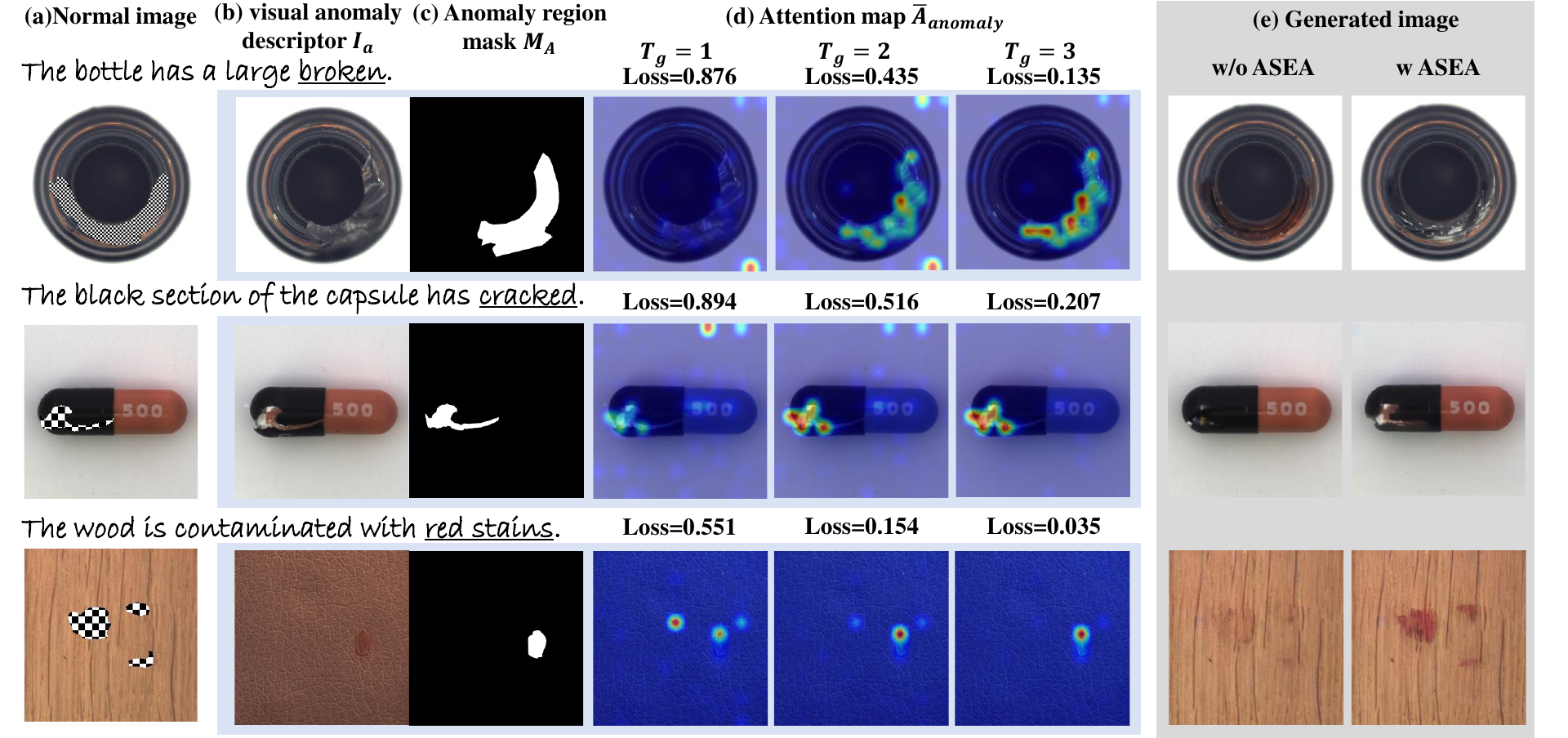}}
\caption{Effect of guidance steps \( T_g \) in ASEA. As \( T_g \) increases, the loss value decreases, where ASEA progressively refines VLM's attention on anomaly regions. The generated images exhibit improved anomaly realism, confirming the effectiveness of ASEA in focusing on specified anomaly regions and producing more visually realistic anomalies.}

\label{fig:ASEA_ablation}
\end{figure}

\subsection{Experimental Results}

\textbf{Comparisons in Anomaly Synthesis.} Table \ref{Quantitative generation results} illustrates the quantitative results for the generation task on the MVTec AD dataset. Our method achieves the highest average IS of 1.84 and the highest average IL of 0.35, surpassing all other methods in both overall quality and generalization. These results demonstrate our model's ability to generate high-quality, diverse anomalies that closely resemble real-world anomalies while maintaining robustness across different anomaly types.

As shown in Figure \ref{supp_ad_1}, our approach can generate realistic anomalies that are contextually consistent with the original images. For instance, in the hazelnut example, the generated anomaly of a cracked and broken shell is highly consistent with the texture and appearance of the original object, while maintaining a precise match with the defined mask area. Similarly, in the capsule and wood examples, our method accurately generates anomalies like incorrect imprints and scratches, showing its capability to faithfully replicate defects.

\textbf{Comparisons in Downstream Tasks.} As shown in Table \ref{anomaly detection small}, our model consistently outperformed other anomaly synthesis methods across most evaluation metrics. By enabling fine-grained controls over the appearance of anomalies, our approach can ensure that the generated samples closely resemble real-world defects. 
This capability is crucial in providing diverse and realistic training data, allowing downstream models to generalize better to real-world anomalies and improve both detection sensitivity and localization accuracy.

\textbf{Qualitative Comparisons.} As shown in Figure \ref{QE}, our approach outperforms existing anomaly synthesis methods by producing anomalies with enhanced realism, closely aligning with real-world defect patterns.
(1) For anomaly synthesis methods, each row showcases anomalies across various object categories, where binary-value masks indicate the position of the generated anomaly regions. 
Our method excels in generating realistic anomalies in regions where other methods fail to capture fine details, resulting in unrealistic anomalies, artifacts (e.g., \textit{DFMGAN} and \textit{AnoDiff}, ``cable''), or misaligned/fail to generate corresponding anomalies within the mask (e.g., \textit{DFMGAN} and \textit{AnoDiff}, ``grid''). 
(2) For image generation methods, from left to right, the columns display a normal image with a target mask, a visual anomaly descriptor to provide visual guidance, and the generated results using different methods (\textit{i.e.}, reference-only mode ControlNet, IP-Adapter, and our method), respectively. Our method generates anomalies that are well-aligned with the visual cues in the descriptor, producing defects with a high degree of realism and relevance. 

\begin{table}[!t]
\centering

\setlength{\tabcolsep}{2.7pt} %
\begin{tabular}{r|cccccc}

\hline
\multicolumn{7}{c}{MVTec AD} \\ \hline
Method & AUC-P & AP-P & F1-P & AUC-I & AP-I & F1-I \\
\hline
DRAEM 
& 92.2 & 54.1 & 53.1 & 94.6& 97.0 & 94.4 \\
PRN 
& 96.9 & 66.2 & 64.7 & 91.6 & 96.6 & 92.4 \\
DFMGAN 
& 90.0 & 62.7 & 62.1 & 87.2 & 94.8 & 94.7 \\
AnoDiff 
& \underline{99.1} & \underline{81.4} & \underline{76.3} & \underline{99.2} & \underline{99.7} & \textbf{98.7} \\
\hline
Ours 
& \textbf{99.5} & \textbf{87.4} & \textbf{81.2} & \textbf{99.3} & \textbf{99.8} & \underline{98.0} \\
\hline
\hline %
\multicolumn{7}{c}{MPDD} \\ \hline
Method & AUC-P & AP-P & F1-P & AUC-I & AP-I & F1-I \\ \hline
DRAEM 
& 85.2 & 36.7 & 38.3 & 87.6 & 73.4 & 67.8\\
PRN
&90.7 & 40.8 & 41.5 & 86.3 & 77.6 & 76.3\\
DFMGAN 
& 92.4 & 41.6 & 45.5 & 92.5 & 78.5 & 78.9\\
AnoDiff
&\underline{95.0} & \underline{47.6} & \underline{49.5} 
& \underline{95.3} & \underline{88.5} & \underline{88.0} \\
\hline
Ours 
&\textbf{96.5} & \textbf{50.6} & \textbf{51.2} 
& \textbf{98.4} & \textbf{90.5} & \textbf{89.1} \\
\hline
\hline %
\multicolumn{7}{c}{VisA} \\ \hline
Method & AUC-P & AP-P & F1-P & AUC-I & AP-I & F1-I \\ \hline
DRAEM 
& 88.8 & 25.4 & 24.6& 84.1& 88.2& 87.4\\
PRN
& 87.0  & 20.8  & 21.5 &80.5& 81.9& 82.3\\
DFMGAN 
& 95.2& 39.9 & 40.2 &87.1 & 85.2 &85.7\\
AnoDiff
& \underline{98.0} & \underline{33.2}   & \underline{37.1} & \underline{96.2}   & \underline{96.9} & \underline{92.6}  
\\
\hline
Ours 
& \textbf{99.0} & \textbf{42.5}  & \textbf{48.0}   & \textbf{96.8}   & \textbf{97.2} & \textbf{93.5}      \\
\hline
    
\end{tabular}
\caption{Results of anomaly detection. ``-P'' denotes the results of anomaly localization (pixel-level) and ``-I'' denote the results of anomaly detection (image-level).
Bold values indicate the best results, and underlined values indicate the second-best results.
}
\label{anomaly detection small}
\end{table}

\begin{table}[t]
\centering

\setlength{\tabcolsep}{3pt} %
\begin{tabular}{@{}ccc|cccc@{}}
\hline
SGA  & CSM & ASEA-\( T_g \) & IS & IL & AUC-P & AUC-I \\
\hline
- & - & -                   & 1.50 & 0.18 & 90.3 & 93.0 \\
\checkmark & - & -          & 1.66 & 0.23 & 92.7 & 95.4 \\
\checkmark & \checkmark & - & 1.69 & 0.25 & 93.1 & 95.8 \\
\checkmark & \checkmark & 1 & 1.70 & 0.25 & 94.1 & 96.8 \\
\checkmark & \checkmark & 2 & 1.78 & 0.28 & 96.2 & 97.9 \\
\checkmark & \checkmark & 3 & \textbf{1.84} & \textbf{0.35} & \textbf{99.5} & \textbf{99.3} \\
\checkmark & \checkmark & 4 & \textbf{1.84} & \underline{0.33} & 99.3 & \textbf{99.3} \\
\checkmark & \checkmark & 5 & \textbf{1.84} & \textbf{0.35} & \underline{99.4} & \underline{99.0} \\
\hline
\end{tabular}

\caption{Ablation Study. All modules could improve the effectiveness of the method. We also study the impact of different guidance steps \( T_g \) in ASEA.
Bold values indicate the best results, and underlined values indicate the second-best results.
}

\label{tab:ablation_AnomalyControl Adapter_ASEA}

\end{table}

\textbf{Effectiveness of ASEA.} As shown in Figure \ref{fig:ASEA_ablation}, ASEA progressively refines VLM's focus on specified anomaly areas, leading to higher anomaly realism for visually realistic and detailed anomalies. By enhancing focus on anomaly-specific regions, ASEA effectively helps to generate high-quality, detailed anomaly images.

\textbf{Ablation Study.} As shown in Table \ref{tab:ablation_AnomalyControl Adapter_ASEA}, it is observed that all modules could improve the effectiveness of the method. 
With only the SGA module, IS and IL increase to 1.66 and 0.23, respectively, while AUC-P and AUC-I reach 92.7 and 95.4. At this stage, the model only uses the visual anomaly descriptor, functioning as an adapter based solely on the image prompt. The addition of the SGA module shows that SGA can strengthen the focus on the anomaly regions in the image.
With the CSM module, IS rises from 1.66 to 1.69, IL from 0.23 to 0.25, and AUC-P and AUC-I reach 93.1 and 95.8. This improvement indicates that the CSM module incorporates the textual anomaly descriptor, integrating it with the visual anomaly descriptor for cross-modal semantic fusion, thereby improving the quality and coherence of anomaly generation. As described in the methods section, the CSM module captures cross-modal semantic features of the anomaly region through multimodal interactions, resulting in anomaly features that align better with the intended descriptions.
With the ASEA module, different guidance steps \( T_g \) demonstrate its impact on model performance. When \( T_g = 3 \), all metrics reach optimal values, providing the best balance between computational efficiency and performance. Increasing \( T_g \) to 4 or 5 shows marginal improvements, indicating that performance gains plateau while computational costs continue to rise.

We provide additional experimental results, analyses, and qualitative examples in the Appendix.

\section{Conclusion}
In this paper, we propose an AnomalyControl framework for controllable anomaly synthesis, which enables the flexible generation of realistic and generalized anomaly samples. AnomalyControl introduces three modules, \textit{i.e.}, Cross-modal Semantic Modeling (CSM), Anomaly-Semantic Enhanced Attention (ASEA), and Semantic Guided Adapter (SGA). To enhance the realism of the synthesized anomalies, CSM aims to extract detailed cross-modal semantic features based on the textual and visual anomaly descriptors, where ASEA could capture fine-grained visual details through multimodal interactions. Then, prompted by precise and controllable instructions, SGA could leverage these enriched semantic signals to guide the diffusion process for high-quality anomaly synthesis. Extensive experiments indicate that AnomalyControl can achieve state-of-the-art results in anomaly synthesis by generating realistic and generalized anomalies based on different control signals and outperforming existing methods for downstream tasks.

\bibliography{aaai2026}
\clearpage
\newpage
\clearpage

\section{Appendix}

\vspace{1em}
\hrule
\vspace{1em}

This appendix consists of:
\begin{itemize}
    \item Section A: The training algorithm of AnomalyControl, detailing the optimization of cross-modal semantic features.
    \item Section B: Additional implementation details, including dataset descriptions, anomaly mask generation via textual inversion, and caching-based feature extraction.
    \item Section C: Quantitative results on the MPDD and VisA datasets, and comprehensive anomaly detection results across multiple benchmarks.
    \item Section D: Qualitative results illustrating anomaly synthesis across diverse categories, generalization to unseen anomalies, and comparison with previous methods.
\end{itemize}

\vspace{1em}
\hrule
\vspace{1em}

\section{A. AnomalyControl Training Algorithm}

We summarize the training process of AnomalyControl in Algorithm~\ref{anomalycontrol_training} for clarity and reproducibility.

\begin{algorithm}[htbp]
\caption{AnomalyControl Training Algorithm}
\label{anomalycontrol_training}
\textbf{Input}: Visual anomaly descriptor $I_a$, textual anomaly descriptor $T_a$, anomaly region mask $M_A$, targeted text prompt $T_t$, time steps $T$; number of training steps $T_{\text{train}}$, guidance steps $T_g$ \\
\textbf{Output}: Trained model parameters for AnomalyControl
\begin{algorithmic}[1]
\FOR{$t_{\text{train}} = 1$ to $T_{\text{train}}$}
    \STATE Sample a batch of images $I$ from training dataset $I_a$
    \STATE Encode images to latent space: $z = \text{Encoder}(I)$
    \STATE Sample noise $\epsilon \sim \mathcal{N}(0, 1)$
    \STATE Sample time step $t \sim \text{Uniform}(1, T)$
    \STATE Add noise to latent variable $z$ to obtain $z_t$
    \STATE Initialize trainable attention guidance variable $e_g \leftarrow 0$
    \STATE Extract text embedding $e_t$ and visual embedding $e_v$ from frozen VLM
    \FOR{$g = 1$ to $T_g$}
        \STATE Generate attention map $A = \text{VLM}_{\text{crossattn}}(e_t, e_v + e_g)$
        \STATE Compute average attention map $\bar{A}_{\text{anomaly}}$ by Eq.~\ref{A_anomaly}
        \STATE Compute energy loss $E(\bar{A}_{\text{anomaly}}, M_A)$ by Eq.~\ref{energy_loss}
        \STATE Update $e_g$ by Eq.~\ref{e_g}
    \ENDFOR
    \STATE Generate cross-modal semantic feature $F_c = \text{VLM}(e_t, e_v + e_g)$
    \STATE Compute loss $\mathcal{L}$ by Eq.~\ref{loss_new}
    \STATE Backpropagate $\mathcal{L}$ and update SGA parameters
\ENDFOR
\STATE \textbf{return} Trained AnomalyControl model parameters
\end{algorithmic}
\end{algorithm}

\section{B. Implementation Details}
\label{Implementation Details}

\subsection{Dataset}
The MVTec AD dataset consists of 5,354 images across 15 categories for industrial anomaly detection tasks. These categories include 10 object categories and 5 texture categories. Each category is represented by defect-free training images and a test set containing both defect-free and defective images. Pixel-level annotations are provided for all anomalies in this dataset. We utilized the captions from MVTec AD Caption \cite{AnoXFusion} for training and testing, which were carefully designed to ensure cross-modal complementarity.

The MPDD dataset includes 1,346 images from 6 types of industrial metal products, each captured under varying lighting conditions and non-uniform backgrounds. This dataset features multiple products in each image, with diverse placement orientations, shooting distances, and positions, making it a challenging dataset for anomaly detection.
The text descriptions provided for each image consist of the product types and the specific anomaly categories identified.

The ViSA dataset comprises 9,621 normal images and 1,200 anomaly images across 12 categories. The dataset includes categories with complex structures, such as PCBs, and categories that require the detection of multiple objects, such as capsules. These characteristics make it a challenging dataset for both anomaly detection and localization tasks, requiring high precision to accurately identify and classify the anomalies.
The text descriptions are directly provided by the dataset.

While some anomalies are difficult to describe, our method considers this limitation by incorporating image reference prompts in addition to text descriptions. 
The image reference prompt provides a visual example of the anomaly type, enabling our model to synthesize realistic anomalies even when textual descriptions are ambiguous.

\subsection{Mask Generation}
Given the limited number of real anomaly masks in the dataset and the lack of diversity in mask data even after augmentation, we adopt a method similar to AnoDiff \cite{AnoDiff} to generate additional anomaly masks by learning the distribution of real anomaly masks. Specifically, we use the textual inversion technique \cite{textual_inversion} to learn a mask embedding \( e_m \), which serves as a text condition for generating diverse anomaly masks.

Initially, the mask embedding \( e_m \) is composed of randomly initialized tokens, which are then optimized by minimizing the following objective function:
\[
e_m^* = \arg \min_{e_m} \mathbb{E}_{z \sim \mathcal{E}(m), \epsilon, t} \left[ \left\| \epsilon - \epsilon_\theta (z_t, t, e_m) \right\|_2^2 \right].
\]
Once trained, the mask embedding is used as a text condition to guide the generation process within a latent diffusion model (LDM) \cite{SD}. We employ classifier-free guidance \cite{CFG} to adjust the influence of the mask embedding, as follows:
\[
\hat{\epsilon}_\theta (x_t | e_m) = \epsilon_\theta (x_t) + s \cdot \left( \epsilon_\theta (x_t, e_m) - \epsilon_\theta (x_t) \right),
\]
where \( s \) is a scaling factor controlling the strength of the text guidance. Following the setup in textual inversion, \( s \) is set to 5 to ensure that the generated masks accurately capture anomaly-specific features and positional diversity.

\subsection{Feature Extraction with Caching in CSM}

In Cross-modal Semantic Modeling (CSM), the Visual-Language Model is frozen, and the trainable attention guidance variable \( e_g \) is used to guide the VLM in generating cross-modal semantic features that focus more on the anomaly region. Since we do not train the VLM, the extracted features remain fixed when the input is fixed. This design brings a potential benefit: once these features are extracted for the first time during training, they can be cached.
As shown in Algorithm \ref{feature_extraction_with_cache}, in subsequent training steps, whenever these features are needed, they can be loaded directly from the cache, avoiding the need for recomputation. This not only eliminates the overhead of recalculating image and text features but also significantly speeds up the training process of AnomalyControl. This approach saves substantial computational resources and time, improving the overall training efficiency.

\begin{algorithm}[tb]
\caption{Feature Extraction with Caching in CSM}
\label{feature_extraction_with_cache}
\textbf{Input}: Visual anomaly descriptor $I_a$, textual anomaly descriptor $T_a$, anomaly region mask $M_A$, cache directory \texttt{cache\_dir} \\
\textbf{Output}: Extracted cross-modal semantic feature $F_c$
\begin{algorithmic}[1]
\STATE \texttt{cache\_file} $\leftarrow$ \texttt{cache\_dir} / $I_a$.pt
\IF{\texttt{cache\_file exists}}
    \STATE $F_c \leftarrow$ \texttt{load}(\texttt{cache\_file})
\ELSE
    \STATE $F_c \leftarrow$ \texttt{ExtractFeatures}($I_a$, $T_a$, $M_A$)
    \STATE \texttt{save}($F_c$, \texttt{cache\_file})
\ENDIF
\STATE \textbf{return} $F_c$
\end{algorithmic}
\end{algorithm}

\subsection{Training Details}

We use the AdamW optimizer \cite{adamW} with a fixed learning rate of 0.0001 and weight decay of 0.01. For classifier-free guidance, text or cross-modal semantic features are independently dropped with a probability of 0.05, and both are dropped simultaneously with a probability of 0.05. Training is conducted on a single A100 GPU, requiring approximately three days for 200K iterations. During inference, a 30-step DDIM sampler is used, with a guidance scale of 7.5, to achieve a balance between fidelity and generalization in the generated anomalies.

\section{C. More Quantitative experiments} %
\label{More Quantitative experiments}

Table \ref{supp_table1} presents the quantitative generation results on the MPDD and ViSA datasets, where the Inception Score (IS) and Intra-cluster Pairwise Learned Perceptual Image Patch Similarity  (IL) are used to evaluate the quality and diversity of the generated anomalies. 
Our method outperforms existing SOTA models (DFMGAN and AnoDiff) in both datasets, demonstrating superior generation quality. Specifically, on the MPDD dataset, our approach achieves the highest IS of 1.71 and IL of 0.47, while on the ViSA dataset, it reaches an IS of 1.65 and IL of 2.12, surpassing the performance of other methods.

\begin{table}[!t]
\centering

\label{tab:results on VisA datasets}
\begin{tabular}{r|cccccc}
\hline
Dataset & 
\multicolumn{2}{c}{DFMGAN} & 
\multicolumn{2}{c}{AnoDiff} & 
\multicolumn{2}{c}{Ours}
\\ \hline
Metrics & IS & IL & IS & IL & IS & IL
\\ \hline  
MPDD  & 1.31 & 0.31 & 1.60 & 0.43 &  \textbf{1.71} & \textbf{0.47} \\
VisA   & 1.26 & 1.12   & 1.45 & 1.73   & \textbf{1.65} &  \textbf{2.12}          
\\ \hline
\end{tabular}
\caption{Generation quantitative results on MPDD and VisA datasets.}
\label{supp_table1}
\end{table}

Table \ref{supp_table_2} further demonstrates our method’s robustness by providing anomaly detection results on the MVTec AD, MPDD, and ViSA datasets. This table includes the anomaly detection results for each anomaly object on these datasets, both at the pixel level (AUC-P, AP-P, F1-P) and image level (AUC-I, AP-I, F1-I). 
Our method consistently delivers high performance across all datasets, with particularly strong results in MVTec AD, where it achieves a mean F1-P of 81.2 and F1-I of 98.0. 
Additionally, our method shows impressive performance on both the MPDD and ViSA datasets, with a mean F1-I of 89.1 on MPDD and 93.5 on ViSA. 
These results underscore the effectiveness of our approach in detecting and localizing anomalies across various datasets and anomaly types. This effectiveness is attributed to the higher quality of the anomalies generated by our method, which provides more realistic and diverse training data for anomaly detection tasks.

\begin{table}[!t]
\centering

\caption{Results of anomaly detection on three datasets. ``-P'' denotes the results of anomaly localization (pixel-level) and ``-I'' denote the results of anomaly detection (image-level).}

\setlength{\tabcolsep}{2.4pt} %
\begin{tabular}{r|cccccc}

\hline
\multicolumn{7}{c}{MVTec AD} \\ \hline
Object & AUC-P & AP-P & F1-P & AUC-I & AP-I & F1-I \\
\hline
capsule & 98.9 & 72.9 & 62.5 & 99.5 & 99.1 & 95.9 \\
bottle & 99.5 & 93.0 & 92.4 & 100 & 100 & 100 \\
carpet & 99.4 & 74.1 & 73.9 & 97.7 & 99.4 & 94.4 \\
leather & 99.8 & 79.9 & 74.1 & 100 & 100 & 99.2 \\
pill & 99.8 & 96.4 & 90.0 & 98.3 & 99.6 & 97.4 \\
transistor & 99.6 & 94.4 & 87.2 & 100 & 100 & 100 \\
tile & 99.5 & 94.5 & 89.0 & 100 & 100 & 100 \\
cable & 99.2 & 87.8 & 86.3 & 99.8 & 99.8 & 98.4 \\
zipper & 99.6 & 86.8 & 78.6 & 100 & 100 & 100 \\
toothbrush & 99.5 & 81.7 & 80.1 & 100 & 100 & 100 \\
metal nut & 99.7 & 98.3 & 93.3 & 99.9 & 100 & 99.2 \\
hazelnut & 99.7 & 96.1 & 89.8 & 99.9 & 100 & 99.0 \\
screw & 99.4 & 86.6 & 80.1 & 95.1 & 99.5 & 91.7 \\
grid & 98.9 & 68.8 & 59.2 & 99.5 & 99.1 & 98.7 \\
wood & 99.8 & 99.3 & 82.2 & 99.7 & 99.9 & 98.8 \\
\hline
MEAN & 99.5 & 87.4 & 81.2 & 99.3 & 99.8 & 98.0 \\
\hline
\hline %
\multicolumn{7}{c}{MPDD} \\ \hline
Object & AUC-P & AP-P & F1-P & AUC-I & AP-I & F1-I \\
bracket black & 95.4 & 30.6 & 35.5 & 98.5 & 82.5 & 80.9 \\
bracket white & 95.2 & 31.7 & 38.5 & 98 & 83.2 & 81.1 \\
bracket brown & 96.2 & 36.7 & 33.3 & 96.6 & 83.4 & 82.8 \\
connector & 95.9 & 64 & 54.5 & 100 & 100 & 100 \\
metal plate & 99.5 & 87.4 & 88.2 & 100 & 100 & 100 \\
tubes & 96.6 & 53.4 & 57.5 & 97.3 & 93.8 & 89.9 \\
\hline
MEAN 
&96.5 & 50.6 & 51.2 
& 98.4 & 90.5 & 89.1 \\
\hline
\hline %
\multicolumn{7}{c}{VisA} \\ \hline
Object & AUC-P & AP-P & F1-P & AUC-I & AP-I & F1-I \\ \hline
pcb1 & 99.5 & 51.2 & 55.4 & 97.8 & 95.5 & 95.1 \\
pcb2 & 100 & 40.4 & 48.9 & 96.8 & 94.3 & 91.2 \\
pcb3 & 99.3 & 47.1 & 52.9 & 98.2 & 95.5 & 95.7 \\
pcb4 & 98 & 34.4 & 42.3 & 97.1 & 100 & 100 \\
macaroni1 & 100 & 34.8 & 30.3 & 96.6 & 100 & 93.7 \\
macaroni2 & 98.9 & 30.1 & 39.7 & 95.5 & 95.1 & 93.8 \\
capsules & 96.9 & 34.2 & 33.2 & 96.7 & 100 & 92.3 \\
candle & 99.3 & 16.3 & 26.4 & 95.2 & 97.9 & 73 \\
cashew & 97.6 & 15.8 & 22.3 & 97.2 & 95.4 & 93.4 \\
chewinggum & 100 & 74 & 79.1 & 97.8 & 97.5 & 100 \\
fryum & 99.3 & 43 & 50.3 & 93.9 & 97.1 & 95.9 \\
pipe fryum & 99.3 & 88.1 & 95.3 & 98.4 & 98.4 & 97.9 \\
\hline
MEAN 
& 99.0 & 42.5  & 48.0   & 96.8   & 97.2 & 93.5      \\

\hline
    
\end{tabular}
\label{supp_table_2}
\end{table}

\section{D. More Qualitative experiments} %
\label{More Qualitative experiments}

\begin{figure*}[!t]
\centerline{\includegraphics[width=1.5\columnwidth]{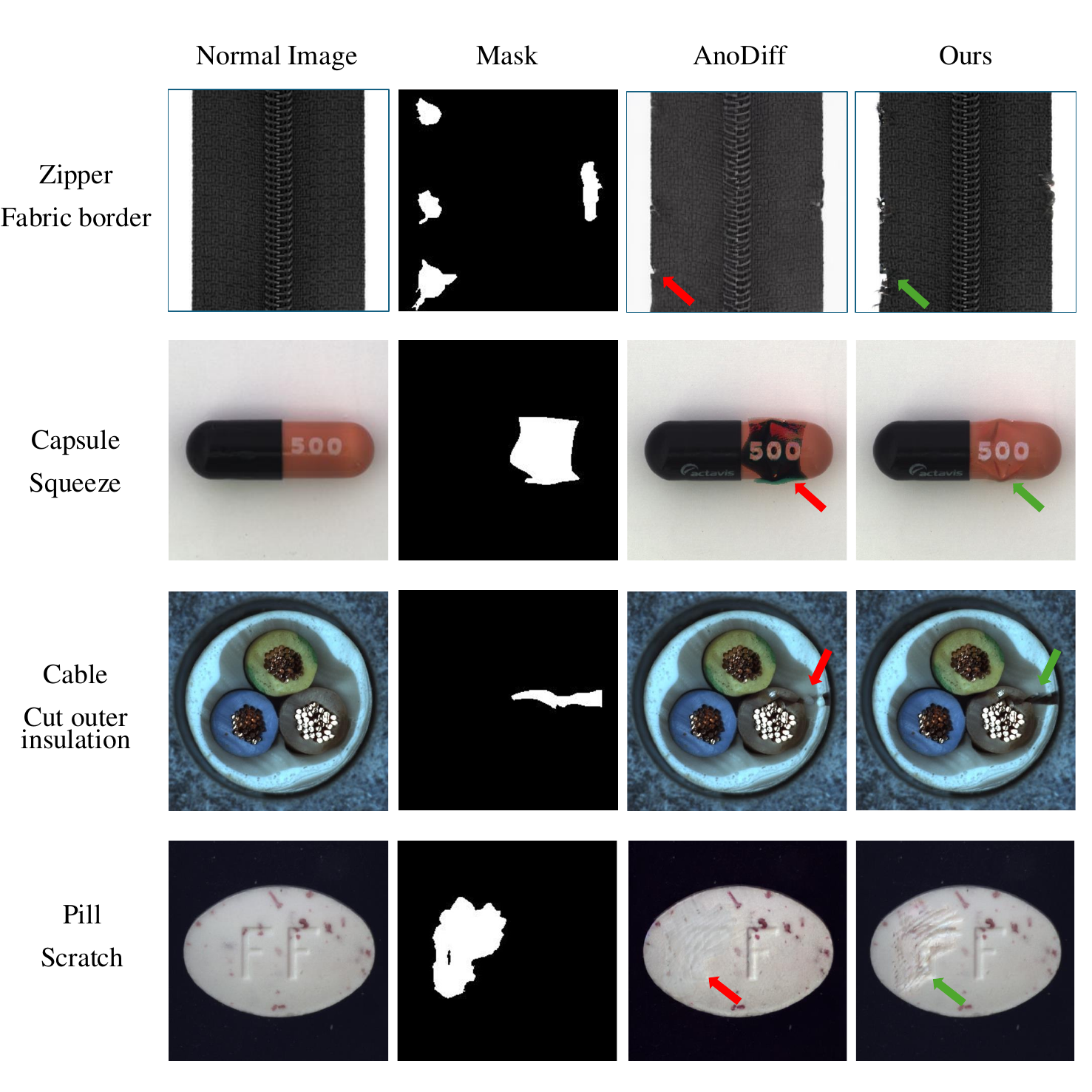}}
\caption{
    Comparison of anomaly synthesis results between AnoDiff and our method. The images show the same normal image (first column) and the same anomaly mask (second column), with the results generated by AnoDiff (third column) and our method (fourth column). 
    In the ``zipper'' example, our method generates an anomaly that better fits the mask and is closer to the real anomaly, even capturing the details of the broken fibers. In the ``capsule'' example, our method successfully generates the squeezed capsule's shape, while AnoDiff fails to synthesize the anomaly. In the ``cable'' example, AnoDiff produces an anomaly with the artifact, while our method generates a more realistic one. In the ``pill'' example, our method's anomaly better matches the mask, whereas AnoDiff produces a blurred anomaly boundary.
}
\label{supp_wood}
\end{figure*}

\subsection{Generalization and Controllability in Anomaly Synthesis}

\begin{figure}[!t]
\centerline{
\includegraphics[width=0.82\columnwidth]{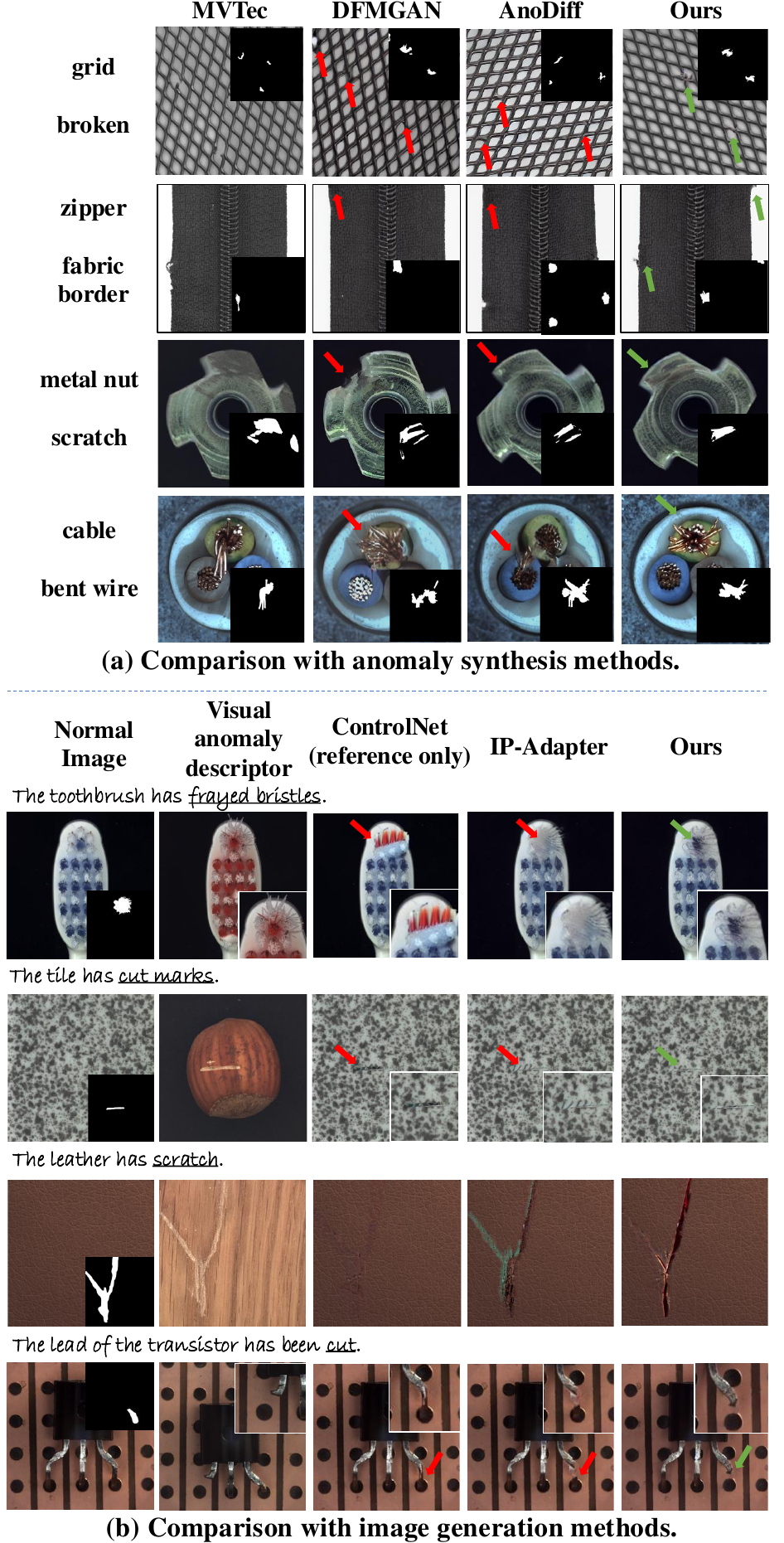}
}
\caption{
Qualitative Evaluation. \textbf{Our method could exhibit large improvements in realism and generalization.}
\textbf{(a) Comparison with anomaly synthesis methods.} Each row showcases anomalies across various object categories, where binary-value masks indicate the position of the generated anomaly regions. 
Our method excels in generating realistic anomalies in regions where other methods fail to capture fine details, resulting in unrealistic anomalies (e.g., \textit{DFMGAN}, ``metal nut''), artifacts (e.g., \textit{DFMGAN} and \textit{AnoDiff}, ``cable''), or misaligned/fail to generate corresponding anomalies within the mask (e.g., \textit{DFMGAN} and \textit{AnoDiff}, ``grid'' and ``zipper''). 
\textbf{(b) Comparison with image generation methods.}
From left to right, the columns display a normal image with a target mask, a visual anomaly descriptor to provide visual guidance, and the generated results using different methods (\textit{i.e.}, reference-only mode ControlNet, IP-Adapter, and our method), respectively. 
Our method generates anomalies that are well-aligned with the visual cues in the descriptor, producing defects with a high degree of realism and relevance. 
}
\label{QE_appendix}
\end{figure}

As shown in Figure \ref{fig1} (right examples) in the main paper and Figure \ref{QE_appendix}(b), our method effectively generalizes to unseen anomaly types using text-image reference prompts for flexible and controllable synthesis.
We further demonstrate the generalization capability of our approach in Figure \ref{supp_wood}, where anomalies are synthesized on different materials (wood and leather) with various anomaly types (such as red stains and holes). These examples highlight our method's flexibility in generating diverse anomalies, even beyond the training distribution. 
The key advantage of our approach is its ability to synthesize diverse and realistic anomalies beyond the training distribution, providing varied training data for detection tasks.

\subsection{Anomaly Synthesis Across Diverse Datasets}
In Figure \ref{supp_mpdd_1} and Figure \ref{supp_visa_1}, we demonstrate additional examples of anomaly synthesis on the MPDD and ViSA datasets, respectively. These results further showcase our model's ability to handle diverse anomaly types and generate realistic anomalies even when faced with different datasets and industrial contexts. 
Even when confronted with varying textures, materials, and anomaly types across these datasets, our method consistently produces high-quality and contextually appropriate anomalies, proving its robustness and adaptability. These examples highlight the model’s generalization capability, ensuring its effectiveness across a wide range of real-world scenarios.

\begin{figure*}[!t]
\centerline{\includegraphics[width=1.9\columnwidth]{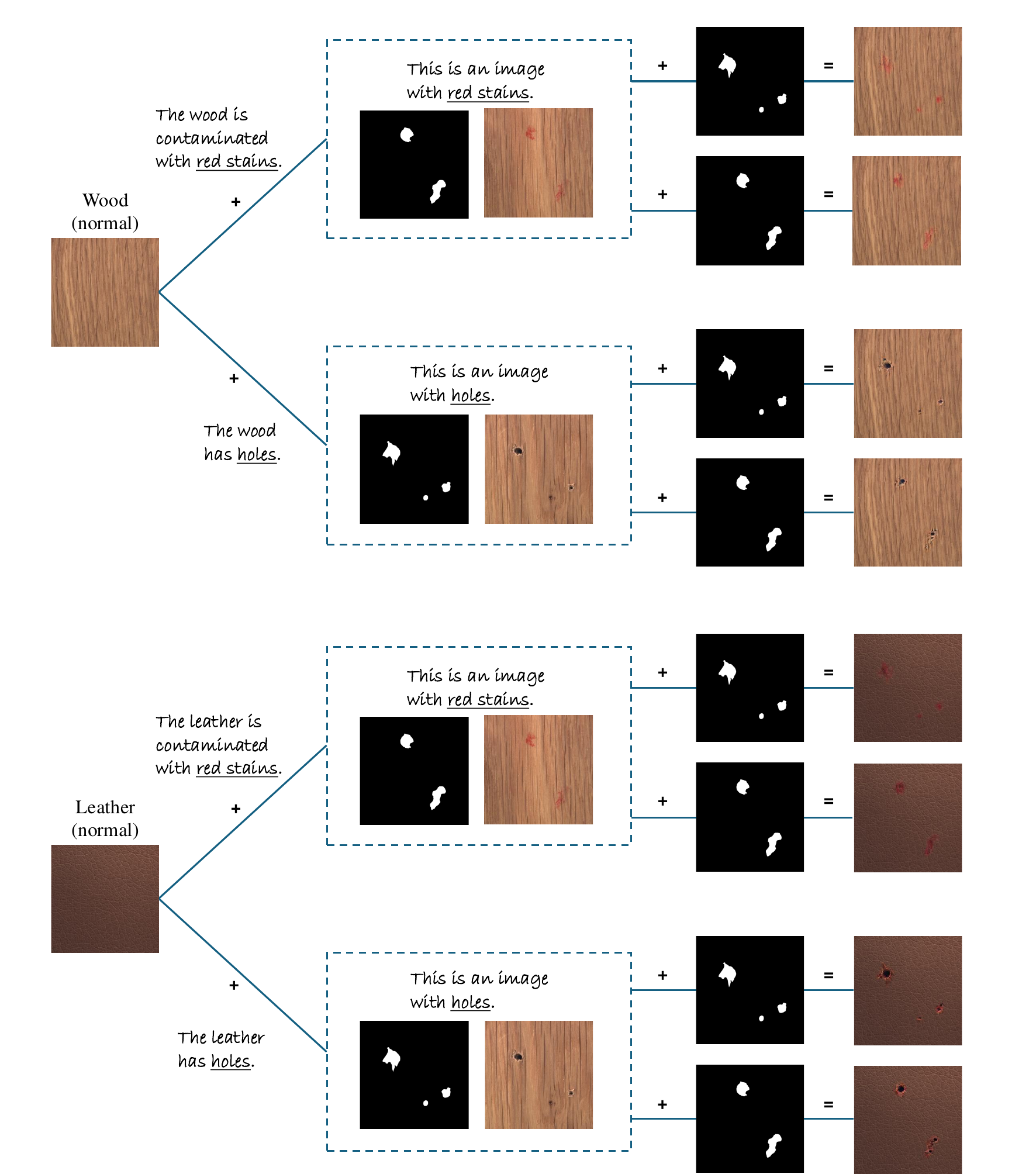}}
\caption{
Controllable anomaly synthesis across different object categories and anomaly types. Given a normal image (wood or leather) and a text-image reference describing the desired anomaly (\textit{e.g.}, ``red stains'' or ``holes''), our method synthesizes corresponding anomalies in the specified masked region. The anomaly type can be flexibly controlled via the reference, and it generalizes well across categories, demonstrating strong generalization capability.
}
\label{supp_wood}
\end{figure*}

\begin{figure*}[!t]
\centerline{\includegraphics[width=2\columnwidth]{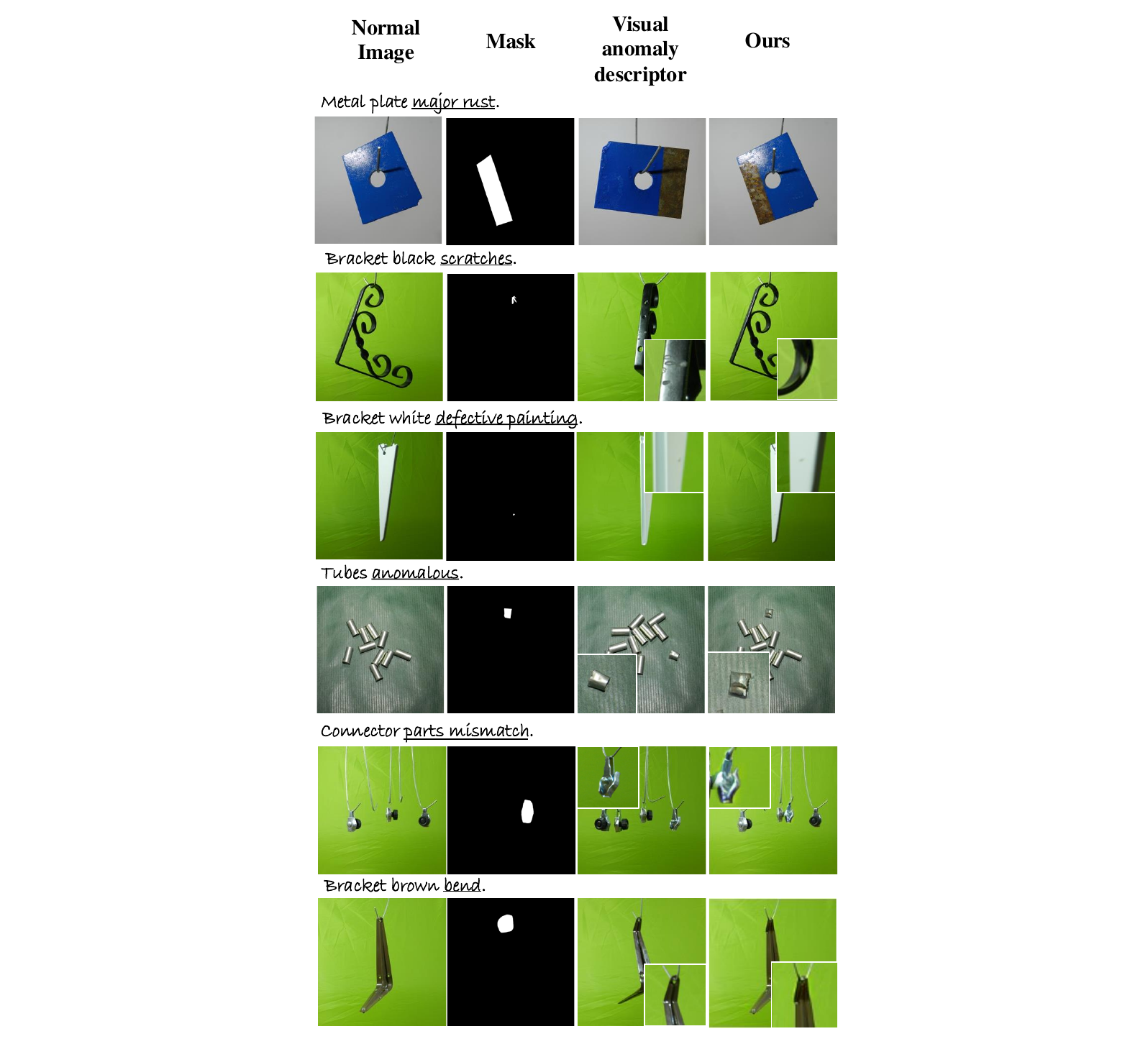}}
\caption{
Display of anomaly synthesis on 6 objects from the MPDD dataset. The first column shows the normal image, the second column represents the mask area where anomalies need to be generated, the third column contains the visual anomaly descriptor providing visual guidance, and the fourth column displays the results generated by our method.  
Unlike simply copying the anomaly shown in the visual anomaly descriptor, our model predicts what the anomaly would look like in the given region indicated by the mask, considering the material and texture of the original normal image. The generated anomalies are consistent with the context of the image and are realistic in their appearance.
Zoom in for better clarity and details.
}
\label{supp_mpdd_1}
\end{figure*}

\begin{figure*}[!t]
\centerline{\includegraphics[width=2\columnwidth]{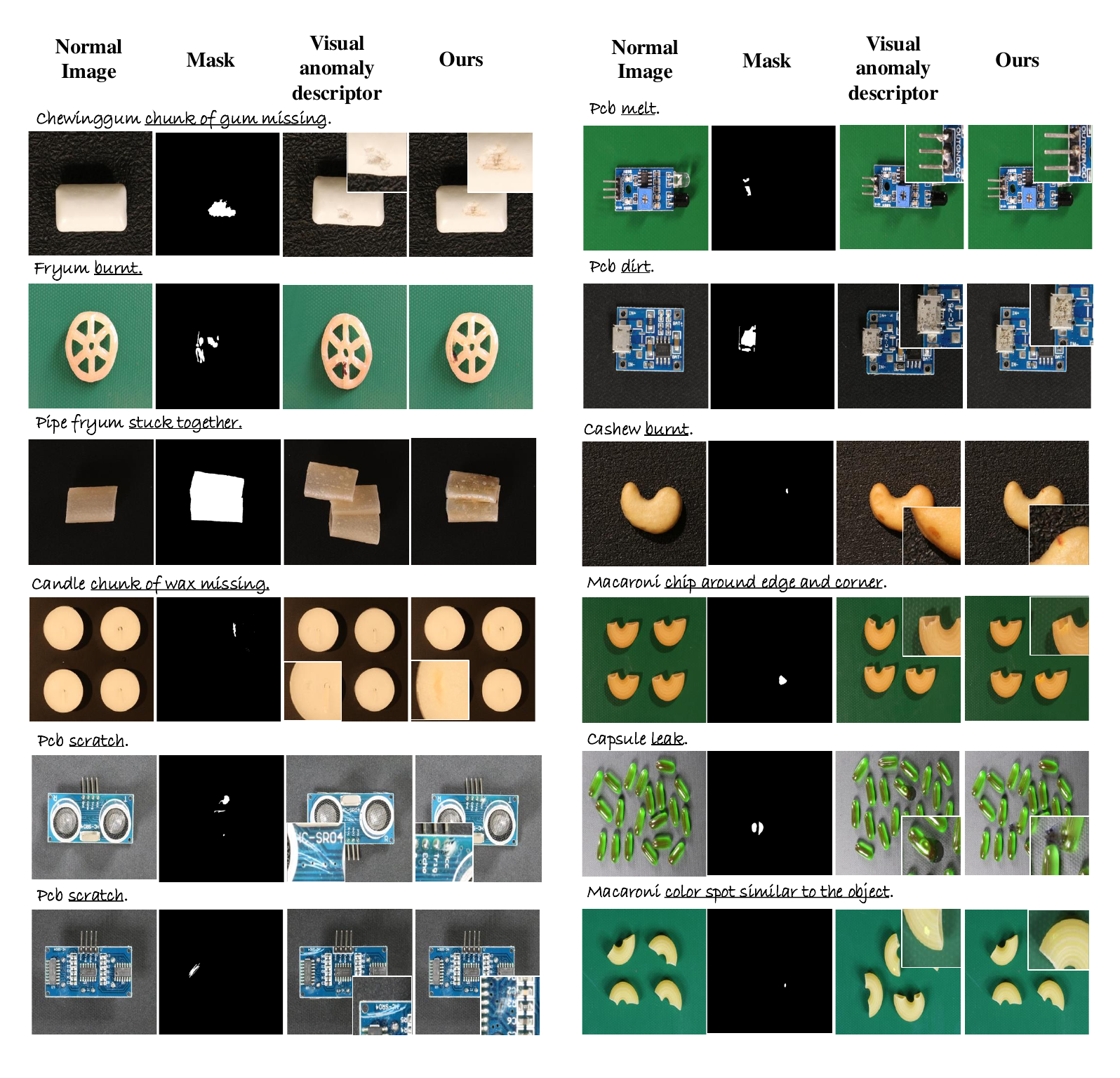}}
\caption{
Display of anomaly synthesis on 12 objects from the ViSA dataset. The first column shows the normal image, the second column represents the mask area where anomalies need to be generated, the third column contains the visual anomaly descriptor providing visual guidance, and the fourth column displays the results generated by our method.  
Unlike simply copying the anomaly shown in the visual anomaly descriptor, our model predicts what the anomaly would look like in the given region indicated by the mask, considering the material and texture of the original normal image. The generated anomalies are consistent with the context of the image and are realistic in their appearance.
Zoom in for better clarity and details.
}
\label{supp_visa_1}
\end{figure*}

\end{document}